\documentclass{article}
\usepackage{iclr2027_conference}
\iclrfinalcopy  %
\usepackage{newtxtext}

\usepackage{amsmath,amsfonts,bm}

\def\eqref#1{equation~\ref{#1}}

\def\1{\bm{1}}

\DeclareMathAlphabet{\mathsfit}{\encodingdefault}{\sfdefault}{m}{sl}
\SetMathAlphabet{\mathsfit}{bold}{\encodingdefault}{\sfdefault}{bx}{n}

\usepackage{hyperref}
\usepackage{url}
\usepackage{booktabs}
\usepackage{graphicx}
\usepackage{amsmath}
\usepackage{amssymb}
\usepackage{multirow}
\usepackage{xcolor}
\usepackage{subcaption}
\usepackage{float}

\usepackage{xspace}
\newcommand{\method}{Fast-VLA\xspace}

\title{\method: You Only Think Once for Robot Manipulation}

\author{
Kehong Gong$^{1}$ \quad
Guanli Hou$^{2}$ \quad
Xiaoxu Zheng$^{1}$ \quad
Qi Wang$^{2}$ \quad
Mingxi Xu$^{2}$ \\
\textbf{Anbang Zhai$^{1}$ \quad
Xiaoyu He$^{2}$ \quad
Mingyuan Zhang$^{2}$\thanks{Corresponding author.} \quad
Michael Bi Mi$^{1}$} \\[4pt]
$^{1}$Huawei Technologies Co., Ltd \qquad
$^{2}$Huawei Central Research Institute
}

\begin{document}

\maketitle
\lhead{Preprint}  %

\raggedbottom

\begin{figure}[H]
\centering
\includegraphics[width=\linewidth]{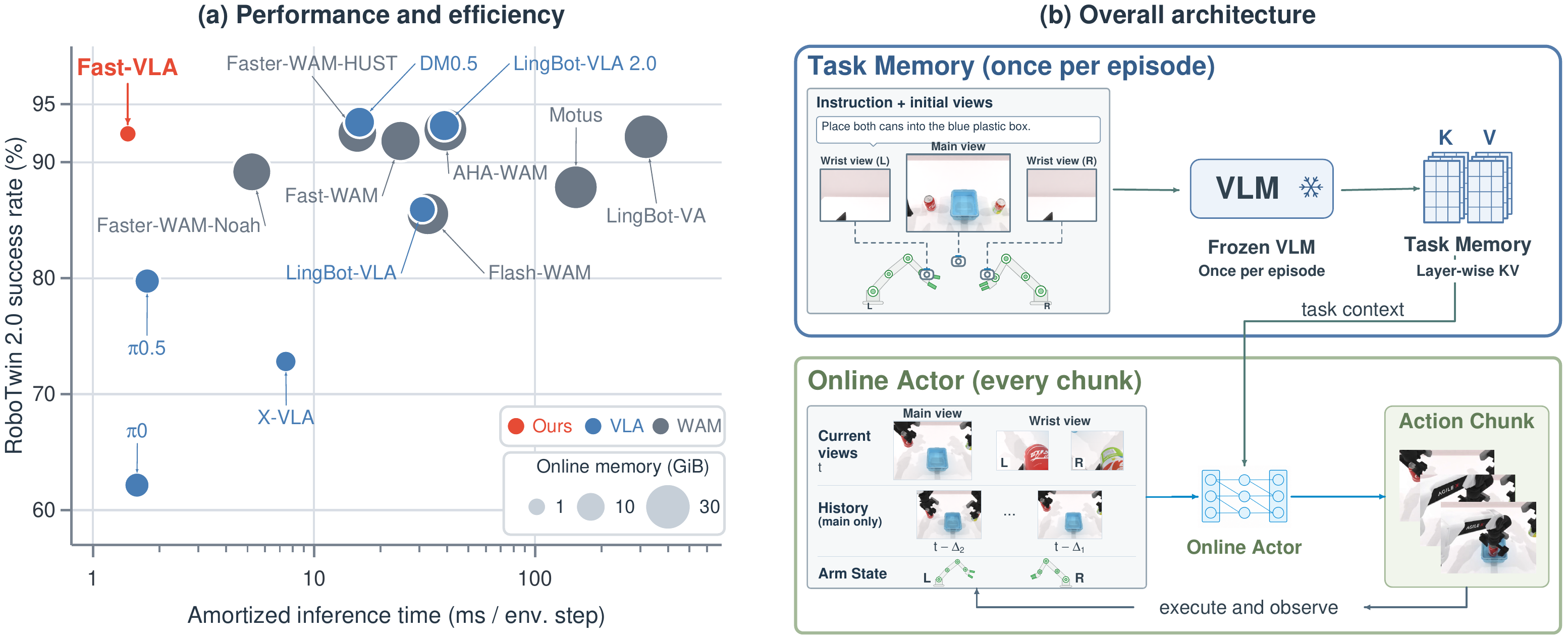}
\caption{\method{} thinks once per episode. (a) Success versus amortized
inference latency on RoboTwin 2.0; bubble area denotes online resident memory
and colour denotes method family. (b) A frozen 4B vision--language model (VLM) encodes the instruction
and initial multi-view frames into fixed, layer-wise KV task memory once at
episode start. At every action chunk, the online actor uses live
observations and robot state to query this memory and predict 50 absolute joint
commands without re-running the VLM.}
\label{fig:teaser}
\end{figure}

\begin{abstract}
Vision--language--action (VLA) policies commonly place a multi-billion-parameter
model inside the feedback loop, repeatedly recomputing task context whenever
they predict a new action chunk. Yet task understanding changes far more slowly
than the robot state. We introduce \method, which invokes a frozen 4B
VLM once per episode to contextualize the instruction and
initial views as fixed, multi-layer key--value task memory. During execution, a
lightweight online actor fuses learned time- and arm-indexed queries with
current visual and proprioceptive feedback to form action queries. These queries retrieve task
context from the fixed memory and predict action chunks without re-running the
foundation model. \method achieves 92.46\% and 92.40\% success on RoboTwin 2.0
Clean and Randomized, and 95.95\% and 79.11\% on LIBERO and
LIBERO-Plus evaluation. Against three continually conditioned baselines on the
same hardware, it reduces RoboTwin amortized policy inference time by
10.9--17.1$\times$ and requires 1.22~GiB of post-prefill online resident
memory. Task-memory swaps further show that the actor relies on episode-matched task
memory. These results show that task context can be established once while
closed-loop control remains responsive throughout execution.
Project page: \url{https://garfield-kh.github.io/FastVLA}.
\end{abstract}

\section{Introduction}
\label{sec:introduction}

VLA models bring the semantic knowledge of large
pretrained VLMs into robot control, enabling a single policy
to follow diverse instructions across objects and scenes
\citep{openvla,pi0,fastwam}. This capability, however, is commonly coupled to
repeated backbone execution: each action chunk re-encodes the current images
and language instruction with a large model. The robot must continuously
observe the environment and revise its actions, but the instruction and overall
task remain largely unchanged. Must the large VLM backbone also be
re-run at every control decision?

When manipulating objects, people typically work out the goal and identify the
relevant objects before acting. As they reach, grasp, and place, they adjust
their movements from what they see and feel without reinterpreting the whole
task before every small action. This suggests two timescales: task
understanding can remain stable, while control must respond continuously to
changes in the scene. This motivates a policy that establishes task context once and then stays
responsive to live observations throughout execution.

Prior work reduces VLA inference cost by changing action representations or
generation procedures \citep{pi0fast,fastwam}. These improvements leave a
separate question: must a large model recompute task context for every action
chunk? Can a VLA think once about the task and still manipulate effectively as
the scene changes throughout an episode?

A fixed memory cannot represent the current world state: objects and arms
move, and the visible scene changes during execution. The central challenge
is to use live observations to query fixed task context at every control
decision. We address this with \method---\emph{You Only Think Once}. A frozen
4B VLM encodes the instruction and initial multi-camera
views into layer-wise key--value task memory at the start of an episode, then
leaves the repeated control loop.

The online actor continually processes current camera views and
proprioception, with short visual history in the RoboTwin configuration.
Query fusion combines learned time- and arm-indexed queries with current arm
state and wrist-view features to form action queries. The action decoder uses
these queries to read visual tokens and task memory, and an action head
predicts an action chunk in one forward pass. The memory remains fixed, but the observations and predicted
actions change at every control decision. The policy therefore remains
closed-loop even though its task memory is fixed.

We evaluate \method on RoboTwin 2.0 \citep{robotwin2}, LIBERO
\citep{libero}, and LIBERO-Plus \citep{liberoplus}. On RoboTwin 2.0, it reaches
92.46\% success in the Clean setting and 92.40\% under domain randomization. \method
obtains 95.95\% on standard LIBERO and 79.11\% on the
LIBERO-Plus protocol. On RoboTwin, same-hardware measurements against
Faster-WAM (HUST), DM0.5, and Fast-WAM show a 10.9--17.1$\times$
reduction in amortized policy inference time after including its one-time
initial call, and its post-prefill online resident memory is 1.22~GiB.
Task-memory swaps cause large performance drops, showing that the actor
depends on episode-matched memory in our evaluation.

Our contributions are threefold:
\begin{itemize}
    \item We introduce an episode-level VLA architecture that represents
    one-time task understanding as fixed, multi-layer memory, moving a frozen
    4B VLM outside the repeated control loop.
    \item We develop an online actor that fuses learned queries with live
    visual and proprioceptive feedback to form action queries. These queries
    retrieve task context across VLM depths, enabling action-chunk prediction
    in one forward pass.
    \item Across RoboTwin 2.0, LIBERO, and LIBERO-Plus, we show that this
    separation preserves competitive closed-loop performance while reducing
    RoboTwin amortized inference cost by 10.9--17.1$\times$; controlled memory
    interventions verify that the actor uses episode-matched context.
\end{itemize}

\section{Related Work}
\label{sec:related-work}

\paragraph{Efficient VLA policies.}
Generalist VLA policies couple pretrained VLM representations
with action generation through autoregressive tokens, flow matching, or
specialized action heads \citep{openvla,pi0}. Recent work reduces the cost of
each policy invocation through parallel continuous-action decoding and action
chunking \citep{openvlaoft}, or through more compact action tokenization
\citep{pi0fast}. These methods primarily optimize action generation while
retaining repeated processing of new observations by the VLM
backbone. Our focus is instead the frequency of backbone invocation: \method
removes the frozen VLM from the control loop after one episode-level prefill.
HyperVLA amortizes task conditioning by encoding the instruction and initial
image into the generated parameters of a compact base policy, which then maps
each current observation to an action~\citep{hypervla}. \method instead keeps
the online actor shared across episodes and retains task context as explicit,
layer-wise key--value task memory; at every control decision, query fusion
combines learned queries with current feedback to form action queries that
retrieve from this fixed memory.

\paragraph{Reusing computation across control steps.}
VLA-Cache reuses representations of visually stable tokens between adjacent
observations while selectively recomputing changed or task-relevant content
\citep{vlacache}. World-action models reduce online cost by omitting test-time
future generation \citep{fastwam}, sparsifying video--action interaction
\citep{fasterwamhust}, or attaching a shallow action module to a deep video
backbone \citep{fasterwamnoah}. These approaches reduce the work performed at
each online call, but retain a large visual or video backbone in the repeated
control path. \method instead fixes task context at episode start and delegates
changing-state estimation entirely to a lightweight online actor.

\begin{figure}[!t]
\centering
\includegraphics[width=\linewidth]{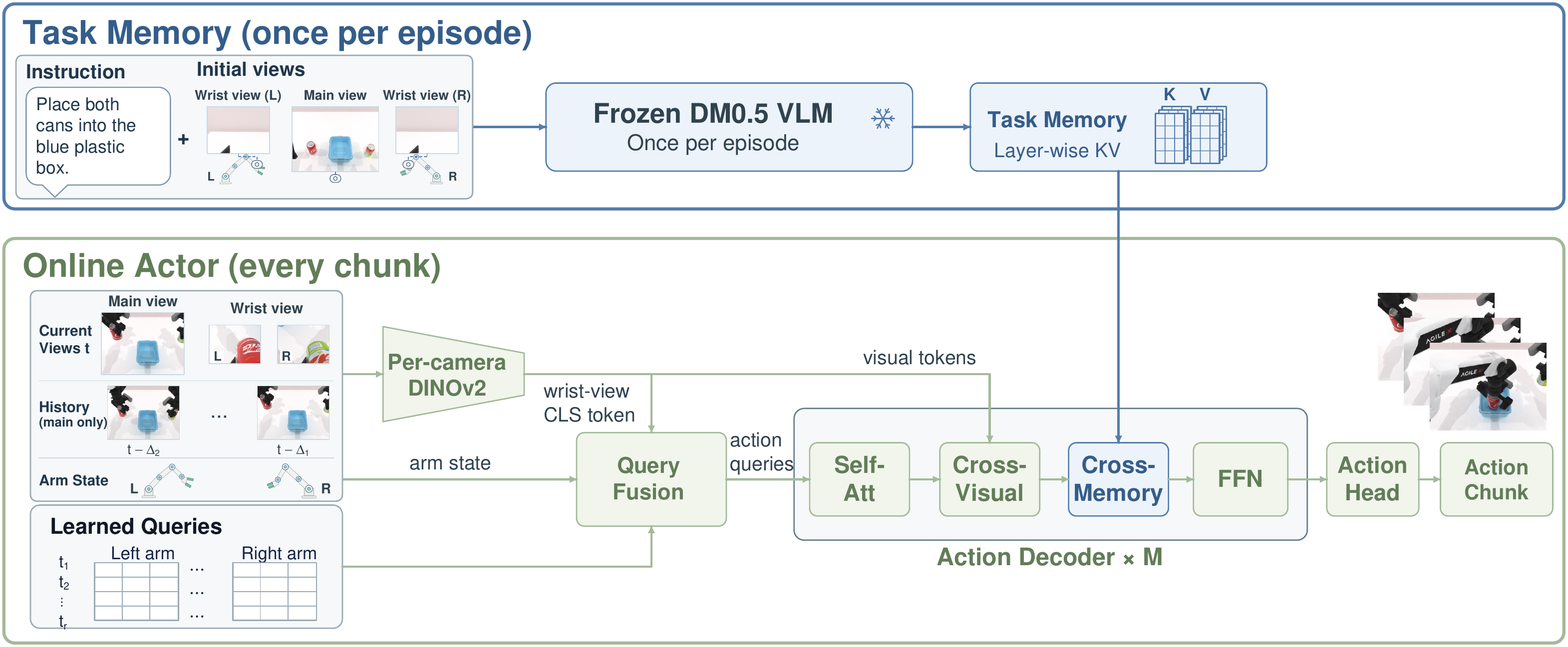}
\caption{Memory-conditioned online control in \method. A frozen VLM constructs
layer-wise KV task memory from the instruction and initial views once per
episode. In the online actor, query fusion combines learned queries with arm
state and wrist-view CLS tokens to form action queries. The action decoder
reads visual tokens from current and historical views and retrieves task
context from the fixed memory. The action head predicts a 50-step bimanual
action chunk.}
\label{fig:architecture}
\end{figure}

\section{Method}
\label{sec:method}

\subsection{Overview}
Figure~\ref{fig:architecture} illustrates the two stages in the RoboTwin
setting. Given an instruction $L$ and initial multi-camera views $I_0$, a
frozen VLM constructs task memory once per episode:
\begin{equation}
M_0=G_{\mathrm{VLM}}(L,I_0).
\end{equation}
The online actor comprises visual encoders, query fusion, an action decoder,
and an action head. At each control decision, its input encoding stage maps
current views $I_t$, proprioceptive state $q_t$, and optional visual history
$H_t$ to online input representations:
\begin{equation}
Z_t=G_{\mathrm{online}}(I_t,H_t,q_t).
\end{equation}
Here $Z_t$ collectively denotes visual tokens and per-arm feedback. The
prediction stage fuses learned queries with this feedback, then uses the action
decoder and action head to read task memory and predict $K$ actions:
\begin{equation}
\hat A_{t:t+K-1}=F_{\mathrm{pred}}(Z_t;M_0).
\end{equation}
Task memory stores episode-level context. At each action-chunk boundary,
the actor combines current visual and proprioceptive observations with this
fixed memory to predict the next action chunk.

\subsection{Episode-level task memory}
\label{sec:memory}
The upper path in Figure~\ref{fig:architecture} uses a frozen DM0.5~\citep{dm05}
4B VLM to contextualize the instruction with the initial
views. We retain the key and value tensors from all 34 Transformer layers:
\begin{equation}
M_0=\{(K_0^\ell,V_0^\ell)\}_{\ell=1}^{34}.
\end{equation}
Retaining all 34 layers allows the actor to retrieve task context at different
VLM depths. The memory remains fixed throughout the episode, and the VLM
leaves the control loop after prefill.

\subsection{Online actor}
\label{sec:actor}
\paragraph{Live inputs.}
In the RoboTwin configuration shown in Figure~\ref{fig:architecture}, the
online actor receives current head and two wrist images, short head-camera
history, and robot state. Each camera has a separate trainable DINOv2~\citep{dinov2} encoder, with
weights shared across frames from that camera. Visual tokens carry learned
slot embeddings and time offsets, distinguishing past from current evidence.
These tokens feed Cross-Visual attention; the CLS token from each current
wrist view separately feeds query fusion.

\paragraph{Query fusion.}
We initialize learned queries for each future time step $\tau$ and arm
$a\in\{\mathrm{left},\mathrm{right}\}$:
\begin{equation}
z_{\tau,a}^{(0)}=e_{\mathrm{time}}(\tau)+e_{\mathrm{arm}}(a),
\qquad \tau\in\{0,\ldots,49\}.
\end{equation}
Query fusion combines these learned queries with each arm's current state and
wrist-view CLS token to form action queries. For arm $a$, we project its
seven-dimensional proprioceptive state and wrist-view CLS token, concatenate
them with its role embedding, and fuse the features through an MLP to obtain
arm feedback $c_t^a$. A gated residual MLP combines each normalized query with
the corresponding feedback and adds the result to the query. The same arm
feedback is reinjected through a layer-specific gated residual before each
subsequent decoder layer. Each action query represents one arm at one future
step and can attend to visual tokens from all cameras. Current feedback
updates the queries at every chunk boundary, allowing the same fixed task
memory to be read differently as execution progresses.

\paragraph{Action decoding.}
Each of the four action decoder layers follows the four stages shown in
Figure~\ref{fig:architecture}. \textsc{Self} attention coordinates queries
across future time steps and the two arms. \textsc{Cross-Visual} attention
reads visual tokens from current views and available history, grounding each
query in visual feedback. \textsc{Cross-Memory} attention then retrieves
relevant task context from the fixed task memory, and the feed-forward network
(FFN) updates the query.
Together, these stages ground each action query in both the current
visual--proprioceptive state and episode-level task context, enabling task-aware
action prediction that remains responsive throughout execution.

Within Cross-Memory attention, a KV memory adapter determines which VLM
depths each decoder layer reads. Decoder
layer $m$ learns scores $s_{m,\ell}$ over the $34$ memory layers, initialized
to favor its depth-aligned layer:
\begin{equation}
\alpha_{m,\ell}=\operatorname{softmax}_{\ell}(s_{m,\ell}).
\end{equation}
After bringing keys into a common rotary-position~\citep{rope} space with
$\mathcal R_{c\leftarrow\ell}$, the mixed memory is
\begin{equation}
(\bar K_m,\bar V_m)
=\sum_{\ell=1}^{34}\alpha_{m,\ell}
\bigl(\mathcal R_{c\leftarrow\ell}(K_0^\ell),V_0^\ell\bigr).
\end{equation}
The queries read $(\mathcal R_{m\leftarrow c}(\bar K_m),\bar V_m)$ with one
grouped-query cross-attention~\citep{gqa} operation. A key-side segment
embedding distinguishes
instruction from initial-image tokens. A linear action head predicts
$\hat A_{t:t+49}\in\mathbb{R}^{50\times14}$ in one forward pass, without
iterative denoising.

For single-arm LIBERO, the same two-stage design uses agent-view and wrist
images, one set of action queries, and a $10\times7$ action chunk. The reported
LIBERO configuration uses current frames without visual history.

\subsection{Training and inference}
We train the online actor, including its visual encoders, with an $\ell_1$
loss on normalized actions:
\begin{equation}
\mathcal L_{\mathrm{act}}=
\frac{1}{KD}\sum_{k=0}^{K-1}\sum_{d=1}^{D}
\left|\hat a_{t+k,d}-a_{t+k,d}\right|.
\end{equation}
The VLM remains frozen. At inference, each episode incurs one VLM prefill and then one actor call per
action chunk. For an episode of $H$ executed steps, we report amortized
latency per environment step as
$(T_{\mathrm{prefill}}+\lceil H/K\rceil T_{\mathrm{actor}})/H$,
using the observed episode lengths.

\paragraph{Implementation details.}
\label{sec:implementation}

We train the online actor with AdamW and a cosine learning-rate schedule.
Table~\ref{tab:hyperparameters} summarizes the main training settings.

\begin{table}[!t]
\centering
\caption{Training settings. Batch size is the effective global batch size;
training compute is reported in accelerator-hours.}
\label{tab:hyperparameters}
\small
\begin{tabular}{lcc}
\toprule
Setting & RoboTwin 2.0 & LIBERO \\
\midrule
Learning rate & $3\times10^{-4}$ & $3\times10^{-4}$ \\
Batch size & 256 & 32 \\
Training steps & 200,000 & 30,000 \\
Training compute (accelerator-hours) & $\approx150$ & $\approx5$ \\
\bottomrule
\end{tabular}
\end{table}

\section{Experiments}
\label{sec:experiments}

\paragraph{Training setup.}
We train separate policies on RoboTwin 2.0's 50-task Clean+Randomized
training set and demonstrations from all 40 tasks across the four LIBERO
suites. The action decoder and head are trained from scratch, while the
visual encoders are initialized from DINOv2 and fine-tuned. Since the
task-memory VLM is frozen, we precompute and store its task-memory KV
caches offline and reuse them throughout actor training, avoiding repeated
VLM forward passes. Full training details are provided in
Table~\ref{tab:hyperparameters} and Appendix~\ref{app:configuration}.

\paragraph{Evaluation protocol.}
We follow the established evaluation protocols of RoboTwin 2.0
\citep{robotwin2}, standard LIBERO \citep{libero}, and
LIBERO-Plus \citep{liberoplus}. On RoboTwin 2.0, we evaluate 100 episodes
per task in each setting (10,000 rollouts). On standard LIBERO, we evaluate
all four 10-task suites with 50 initial states per task (2,000 rollouts).
We also evaluate on all 10,030 LIBERO-Plus perturbation instances.
Baseline success rates come from published reports, benchmark compilations,
or our evaluations of released checkpoints, as detailed in
Appendix~\ref{app:evaluation-details}.

\paragraph{Cost measurement.}
Inference costs are measured separately under a common profiling protocol on identical hardware.
Across Tables~\ref{tab:robotwin} and~\ref{tab:libero}, Pred./exec. denotes
actions predicted per call / actions executed before replanning; ms/chunk
is the wall time of one action-generation call, and amortized ms/step
follows the execution schedule and includes any once-per-episode setup.
We report online resident memory separately from the initialization peak
for \method. Appendix~\ref{app:measurement-protocol} provides the
measurement details.

\subsection{RoboTwin 2.0: performance and inference cost}
\label{sec:efficiency}
\begin{table}[!t]
\centering
\caption{RoboTwin 2.0 success rates (\%) and inference costs. Action PT denotes pretraining of action-generating weights on external large-scale action or latent-action data before benchmark adaptation; VLM or visual-backbone pretraining alone does not count.}
\label{tab:robotwin}
\fontsize{7}{8}\selectfont
\setlength{\tabcolsep}{3pt}
\renewcommand{\arraystretch}{1.0}
\begin{tabular}{@{}clccrrr|rrr@{}}
\toprule
Type & Method & \shortstack{Action\\PT} & \shortstack{Pred./\\exec.} & ms/chunk $\downarrow$ & ms/step $\downarrow$ & GiB $\downarrow$ & Clean $\uparrow$ & Rand. $\uparrow$ & Mean $\uparrow$ \\
\midrule
\multirow{7}{*}{VLA} & $\pi_0$~\citep{pi0} & $\checkmark$ & 50/50 & 72.38 & 1.580 & 6.26  & 65.92$^\flat$ & 58.40$^\flat$ & 62.16 \\
 & $\pi_{0.5}$~\citep{pi05} & $\checkmark$ & 50/50 & 80.51 & 1.758 & 6.47  & 82.74$^\flat$ & 76.76$^\flat$ & 79.75 \\
 & X-VLA~\citep{xvla} & -- & 30/30 & 211.03 & 7.452 & 3.29  & 72.90$^\flat$ & 72.80$^\flat$ & 72.85 \\
 & LingBot-VLA~\citep{lingbotvla} & $\checkmark$ & 50/25 & 740.35 & 30.919 & 7.99  & 86.50 & 85.34 & 85.92 \\
 & Qwen-VLA~\citep{qwenvla} & $\checkmark$ & -- & -- & -- & -- & 86.10 & 87.20 & 86.65 \\
 & LingBot-VLA 2.0~\citep{lingbotvlav2} & $\checkmark$ & 50/25 & 931.29 & 38.892 & 11.98  & 93.52 & 92.80 & 93.16 \\
 & DM0.5~\citep{dm05} & $\checkmark$ & 50/50 & 736.10 & 16.069 & 11.68  & \textbf{93.60} & \textbf{93.30} & \textbf{93.45} \\
\midrule
\multirow{7}{*}{WAM} & Motus~\citep{motus} & $\checkmark$ & 16/16 & 2378.24 & 152.965 & 28.17  & 88.66$^\flat$ & 87.02$^\flat$ & 87.84 \\
 & LingBot-VA~\citep{lingbotva} & $\checkmark$ & 32/32 & 9553.12 & 317.765 & 29.59  & 92.90$^\flat$ & 91.50$^\flat$ & 92.20 \\
 & Fast-WAM~\citep{fastwam} & $\times$ & 32/24 & 564.80 & 24.624 & 23.13  & 91.88 & 91.78 & 91.83 \\
 & Flash-WAM~\citep{flashwam} & $\times$ & 32/32 & 985.10 & 32.767 & 26.26  & 88.42 & 82.66 & 85.54 \\
 & AHA-WAM~\citep{ahawam} & $\times$ & 16/16$^\dagger$ & 553.27$^\dagger$ & 39.291 & 28.21  & 93.40 & 92.20 & 92.80 \\
 & Faster-WAM (Noah)$^\sharp$~\citep{fasterwamnoah} & $\times$ & 32/24 & 119.94 & 5.229 & 21.34  & 89.70 & 88.64 & 89.17 \\
 & Faster-WAM (HUST)~\citep{fasterwamhust} & $\times$ & 32/28 & 416.12 & 15.704 & 22.27  & 92.78 & 92.26 & 92.52 \\
\midrule
Ours & \method & $\times$ & 50/50 & \textbf{44.87}$^\P$ & \textbf{1.437}$^\P$ & \textbf{1.22}$^\P$  & 92.46 & 92.40 & 92.43 \\
\bottomrule
\end{tabular}
\vspace{2pt}
\parbox{\linewidth}{\fontsize{8}{9}\selectfont \textbf{Notes.} ``--'' denotes unavailable data; $^\flat$ marks a benchmark compilation,
and $^\sharp$ our implementation used for cost profiling when no official
one is available. $^\dagger$ The action call excludes a 111.40-ms video prefill performed every two chunks; amortized ms/step includes it. $^\P$ The first \method chunk takes 153.74\,ms (108.87-ms VLM prefill + 44.87-ms actor); subsequent chunks take 44.87\,ms, yielding 1.437 amortized ms/step, with a mean RoboTwin episode length of 237.3 steps. Online resident memory after VLM offloading is 1.22\,GiB; peak memory during prefill is 12.11\,GiB.}
\end{table}
\method achieves 92.46\% Clean and 92.40\% Randomized success
(92.43\% mean) on RoboTwin 2.0 (Table~\ref{tab:robotwin}). Among baselines
in the table with at least 90\% mean success and measured inference costs,
\method provides at least $10.9\times$ faster amortized inference.
Relative to DM0.5, its mean success is 1.02 percentage points lower with
$11.2\times$ faster amortized inference at the same replanning interval of
50 environment steps. Compared with Faster-WAM (HUST), \method scores
0.09 points lower with a $10.9\times$ speedup; it exceeds Fast-WAM by
0.60 points with a $17.1\times$ speedup. Faster-WAM (Noah), at 89.17\%
mean success, has the lowest measured amortized latency among the WAM
baselines; \method exceeds its mean success by 3.26 points with a
$3.6\times$ speedup.

\subsection{LIBERO and LIBERO-Plus}
\begin{table}[!t]
\centering
\caption{Standard LIBERO and LIBERO-Plus success rates (\%) with inference cost.}
\label{tab:libero}
\fontsize{7}{8}\selectfont
\setlength{\tabcolsep}{3pt}
\renewcommand{\arraystretch}{1.0}
\begin{tabular}{@{}clccrrr|rr@{}}
\toprule
Type & Method & \shortstack{Action\\PT} & \shortstack{Pred./\\exec.} & ms/chunk $\downarrow$ & \shortstack{Amort.\\ms/step $\downarrow$} & GiB $\downarrow$ & \shortstack{LIBERO\\Std. $\uparrow$} & \shortstack{LIBERO\\Plus $\uparrow$} \\
\midrule
\multirow{8}{*}{VLA} & $\pi_0$~\citep{pi0} & $\checkmark$ & 50/5 & 81.96 & 16.577 & 6.26 & 94.10 & 53.60 \\
 & $\pi_0$-FAST~\citep{pi0fast} & $\checkmark$ & 10/5 & 353.06 & 71.885 & 5.57 & 85.50$^\flat$ & 61.60 \\
 & OpenVLA-OFT~\citep{openvlaoft} & $\checkmark$ & 8/8 & 107.00 & 13.685 & 14.41 & 97.10$^\flat$ & 69.60 \\
 & $\pi_{0.5}$~\citep{pi05} & $\checkmark$ & 10/5 & 86.35 & 17.466 & 6.47 & 96.90 & 85.70 \\
 & UniVLA~\citep{univla} & $\checkmark$ & 12/1 & 241.96 & 241.960 & 14.15 & 95.50$^\flat$ & 43.90 \\
 & X-VLA~\citep{xvla} & -- & 30/30 & 199.40 & 7.220 & 3.29 & 98.10 & 72.56$^\parallel$ \\
 & HyperVLA$^{\dagger}$~\citep{hypervla} & $\checkmark$ & 4/1 & \textbf{9.59} & 16.385 & 1.71 & 89.00 & -- \\
 & DM0.5~\citep{dm05} & $\checkmark$ & 10/10 & 701.56 & 71.780 & 11.58 & 98.50$^\parallel$ & \textbf{85.90}$^\parallel$ \\
\midrule
\multirow{6}{*}{WAM} & Motus~\citep{motus} & $\checkmark$ & -- & -- & -- & -- & 97.70 & -- \\
 & LingBot-VA~\citep{lingbotva} & $\checkmark$ & 16/16 & 6913.53 & 450.416 & 22.96 & 98.50 & -- \\
 & Fast-WAM~\citep{fastwam} & $\times$ & 32/10 & 544.50 & 55.710 & 23.13 & 97.60 & 49.14 \\
 & Joint-WAM~\citep{fasterwamhust} & $\times$ & 32/10 & 916.81 & 93.802 & 23.13 & -- & 66.27 \\
 & Faster-WAM (Noah)$^\sharp$~\citep{fasterwamnoah} & $\times$ & 32/10 & 118.85 & 12.160 & 21.35 & 98.50 & 75.00 \\
 & Faster-WAM (HUST)~\citep{fasterwamhust} & $\times$ & 32/10 & 824.43 & 84.350 & 22.27 & \textbf{99.00} & 73.57 \\
\midrule
Ours & \method & $\times$ & 10/10 & 34.25$^\P$ & \textbf{3.975}$^\P$ & \textbf{0.86}$^\P$ & 95.95 & 79.11 \\
\bottomrule
\end{tabular}
\vspace{2pt}
\parbox{\linewidth}{\fontsize{8}{9}\selectfont \textbf{Notes.} Symbols follow Table~\ref{tab:robotwin}; $^\parallel$ marks our evaluation of a released checkpoint. $^\dagger$ HyperVLA\textquotesingle s 89.00\% success rate is taken from the original paper~\citep{hypervla}, using OXE pretraining followed by LIBERO fine-tuning. Its inference costs are measured separately using our implementation on the same hardware. HyperVLA runs its hypernetwork once at episode start (1168.01\,ms) and the generated base policy at every step (9.59\,ms), yielding 16.385 amortized ms/step. $^\P$ The first \method chunk takes 122.63\,ms (88.38-ms VLM prefill + 34.25-ms actor); subsequent chunks take 34.25\,ms, yielding 3.975 amortized ms/step, with a mean LIBERO episode length of 171.9 steps. Online resident memory after VLM offloading is 0.86\,GiB; peak memory during prefill is 11.74\,GiB.}
\end{table}
\method achieves 95.95\% on standard LIBERO and 79.11\% on
LIBERO-Plus (Table~\ref{tab:libero}). Compared with Faster-WAM (Noah)
and Faster-WAM (HUST), it achieves 4.11 and 5.54 percentage points higher
LIBERO-Plus success with $3.1\times$ and $21.2\times$ faster amortized
inference, respectively, although its standard LIBERO success is lower.
All three methods replan every 10 environment steps. On LIBERO-Plus,
\method trails $\pi_{0.5}$ and DM0.5 by 6.59 and 6.79 percentage points,
respectively. Both baselines use pretrained action-generating weights,
whereas our action decoder and head are trained from scratch on LIBERO
demonstrations.

\subsection{Qualitative rollouts and live visual grounding}
\begin{figure}[!t]
\centering
\begin{minipage}[c]{0.16\linewidth}
\raggedright\small
\textbf{RoboTwin 2.0}\\
Stack two\\blocks
\end{minipage}\hfill
\begin{minipage}[c]{0.825\linewidth}
\includegraphics[width=\linewidth]{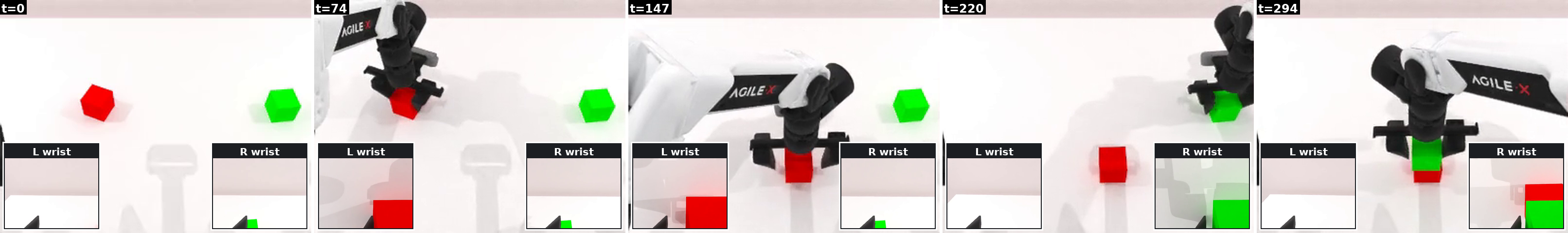}
\end{minipage}
\par\smallskip
\begin{minipage}[c]{0.16\linewidth}
\raggedright\small
\textbf{LIBERO-10}\\
Bowl into\\drawer; close
\end{minipage}\hfill
\begin{minipage}[c]{0.825\linewidth}
\includegraphics[width=\linewidth]{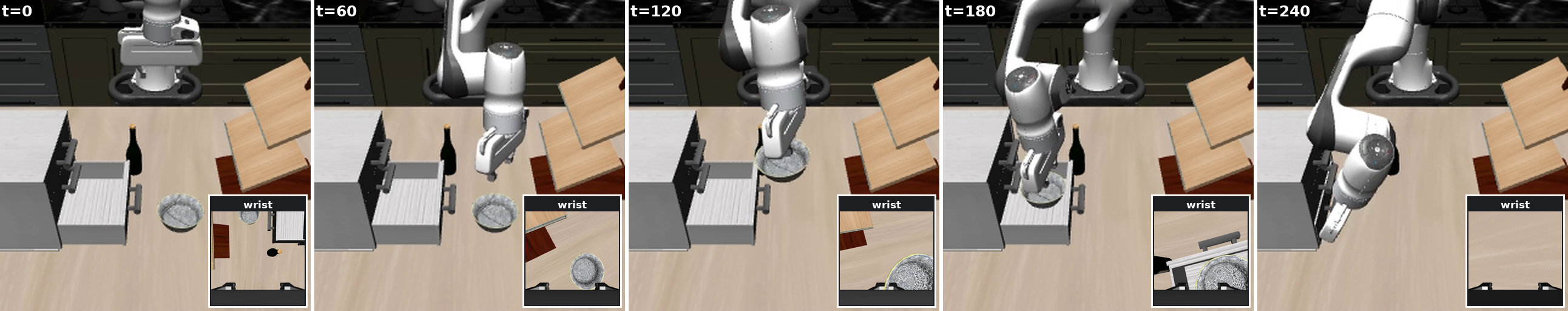}
\end{minipage}
\par\smallskip
\begin{minipage}[c]{0.16\linewidth}
\raggedright\small
\textbf{LIBERO-Plus}\\
Same task;\\sensor noise
\end{minipage}\hfill
\begin{minipage}[c]{0.825\linewidth}
\includegraphics[width=\linewidth]{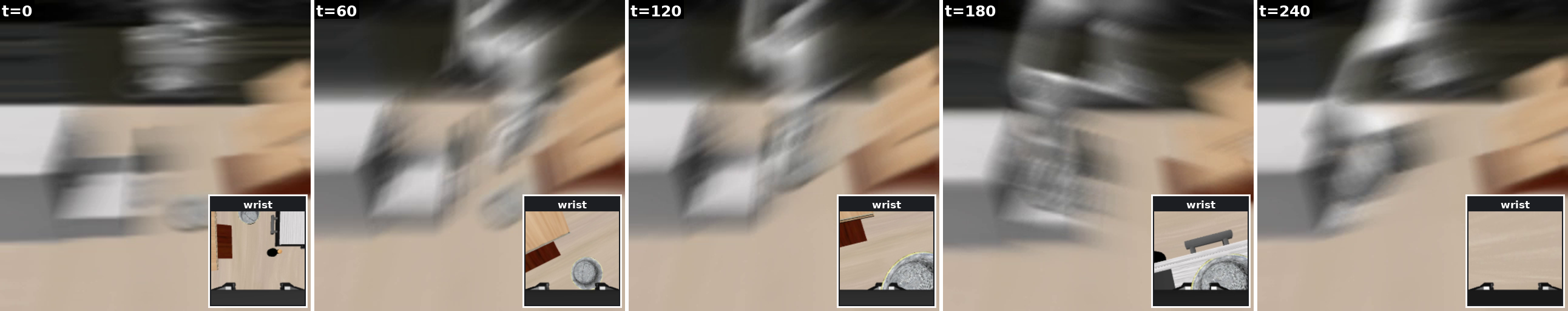}
\end{minipage}
\caption{Successful rollouts with fixed task memory. Each row shows five
frames sampled at equal intervals from one episode; corner labels indicate
frame indices. Wrist views are inset below the third-person view
(two for RoboTwin and one for LIBERO). Task memory is constructed once at
episode start and reused throughout each rollout. In the LIBERO-Plus row,
sensor noise corrupts the third-person view while the wrist view remains
unperturbed; memory is built from the perturbed initial observation.}
\label{fig:rollout-examples}
\end{figure}

\begin{figure}[!t]
\centering
\includegraphics[width=\linewidth]{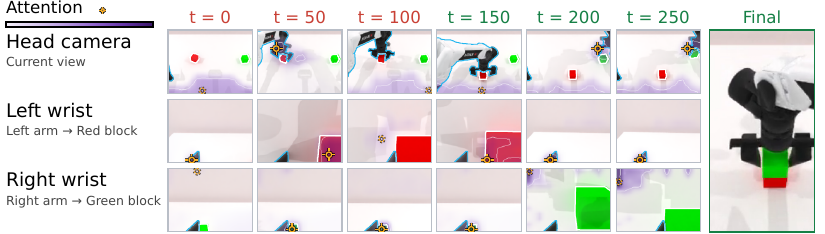}
\caption{Attention follows the active sub-goal. A successful 300-step
\texttt{stack\_blocks\_two} rollout, shown in successive 50-step action chunks.
As control passes from the left arm to the right, head-camera attention shifts
from the red block to the green block. Each wrist focuses on the block handled
by its own arm, with the right wrist attending to the green block before the
second sub-goal begins. Crosshairs mark attention peaks, and enrichment above $1$ indicates
preferential attention to the corresponding block; the final open-loop chunk
has no new attention map.}
\label{fig:attn-flow}
\end{figure}

Figure~\ref{fig:rollout-examples} shows successful rollouts on RoboTwin 2.0
and LIBERO-10, plus the same drawer task under LIBERO-Plus sensor noise.
The noisy episode succeeds despite severe third-person corruption, with an
unperturbed wrist view available to the actor. In all three episodes, task
memory is built once from the initial observations and reused throughout
execution. Appendix~\ref{app:qualitative} shows five additional perturbation
types and further clean rollouts.

\label{sec:live-visual-grounding}
Figure~\ref{fig:attn-flow} visualizes how the actor uses current visual
feedback during a successful \texttt{stack\_blocks\_two} episode. As control
passes from the left arm to the right, head-camera attention shifts from the
red block to the green block, while each wrist focuses on the object relevant
to its arm. The right wrist attends to the green block before the second
sub-goal begins. Together, these shifts show that the actor updates its visual
focus as the active sub-goal changes.

\subsection{Ablations and Analysis}
\label{sec:ablations}
We examine whether the actor relies on task memory and how memory
construction, online observations, refresh frequency, and encoder choice
affect performance. We further analyze success across task horizons to
assess once-per-episode memory reuse on longer tasks.

\paragraph{Task-memory dependence.}
\label{sec:cache-swap}
To test whether the actor actively uses task memory, we alter the task-memory
content at test time while keeping the actor weights fixed without retraining.
\begin{table}[!t]
\centering
\begin{minipage}[t]{0.49\linewidth}
\centering
\caption{RoboTwin 2.0 task-memory swaps (\%). Test-time swaps keep the actor weights fixed, \mbox{without any additional training}.}
\label{tab:cache-swap}
\footnotesize
\setlength{\tabcolsep}{3pt}
\begin{tabular}{@{}p{0.57\linewidth}rr@{}}
\toprule
Memory condition & Clean & Rand. \\
\midrule
Swapped language segment & 35.18 & 29.66 \\
Same task, different episode & 6.26 & 8.36 \\
Swapped initial-view segment & 2.64 & 2.62 \\
Different task & 0.62 & 0.52 \\
\midrule
Original memory (Table~\ref{tab:robotwin}) & 92.46 & 92.40 \\
\bottomrule
\end{tabular}
\end{minipage}\hfill
\begin{minipage}[t]{0.49\linewidth}
\centering
\caption{Task-memory construction on LIBERO (\%). The actor is retrained separately for each task-memory construction variant.}
\label{tab:memory-ablation}
\scriptsize
\setlength{\tabcolsep}{2pt}
\begin{tabular}{@{}p{0.27\linewidth}p{0.34\linewidth}rr@{}}
\toprule
Condition & Memory & Std. $\uparrow$ & Plus $\uparrow$ \\
\midrule
\raggedright No task memory & \raggedright Constant zero KV & 69.90 & 53.69 \\
\raggedright Text-token KV only & \raggedright Image-conditioned text KV & 93.20 & 73.84 \\
\raggedright Last layer only & \raggedright Final-layer instruction + views KV & 94.70 & 73.70 \\
\midrule
Full (Ours) & \raggedright 34-layer instruction + views KV & \textbf{95.95} & \textbf{79.11} \\
\bottomrule
\end{tabular}
\end{minipage}
\end{table}
Swapping the language segment yields $35.18\%$/$29.66\%$ success, while replacing
the initial-view segment causes a larger collapse to $2.64\%$/$2.62\%$.
Memory from another episode of the same task reaches only $6.26\%$/$8.36\%$.
The actor relies on task memory matched to the current episode to guide action prediction.

\paragraph{Memory construction.}
\label{sec:memory-ablation}
Building on the memory-swap results, Table~\ref{tab:memory-ablation}
examines how memory construction affects task performance and robustness.
Removing task memory reduces success to $69.90$/$53.69$ on standard
LIBERO/LIBERO-Plus, highlighting its importance for task execution and
robustness. Retaining only text-token KV from image-conditioned memory
achieves $93.20$/$73.84$, indicating that these tokens preserve useful task
context. Last-layer memory achieves $94.70$/$73.70$, improving standard LIBERO
success over text-token-only memory with a similar LIBERO-Plus success rate.
Full memory achieves the highest success on both benchmarks
($95.95$/$79.11$), supporting the use of instruction and visual KV
across multiple VLM layers.

\paragraph{Online observations.}
\label{sec:input-ablation}
Table~\ref{tab:input-ablation} shows the importance of current wrist vision:
a wrist-only policy reaches $87.10$/$68.72$ on standard LIBERO/LIBERO-Plus,
compared with $70.80$/$30.80$ for the head-only policy. Without wrist vision,
success is $65.60$/$33.49$; without proprioception, it is $94.30$/$72.41$.
These results identify wrist vision as a key source of closed-loop feedback.
\begin{table}[!t]
\centering
\begin{minipage}[t]{0.49\linewidth}
\centering
\caption{Online-input ablations on LIBERO (\%). Each variant is retrained with the indicated \mbox{combination} of online inputs.}
\label{tab:input-ablation}
\footnotesize
\setlength{\tabcolsep}{2pt}
\begin{tabular}{@{}lcccrr@{}}
\toprule
Condition & Head & Wrist & State & Std. $\uparrow$ & Plus $\uparrow$ \\
\midrule
No head view & $\times$ & $\checkmark$ & $\checkmark$ & 92.70 & 74.54 \\
No wrist vision & $\checkmark$ & $\times$ & $\checkmark$ & 65.60 & 33.49 \\
No state & $\checkmark$ & $\checkmark$ & $\times$ & 94.30 & 72.41 \\
Head only & $\checkmark$ & $\times$ & $\times$ & 70.80 & 30.80 \\
Wrist only & $\times$ & $\checkmark$ & $\times$ & 87.10 & 68.72 \\
\midrule
Full (Ours) & $\checkmark$ & $\checkmark$ & $\checkmark$ & \textbf{95.95} & \textbf{79.11} \\
\bottomrule
\end{tabular}
\end{minipage}\hfill
\begin{minipage}[t]{0.49\linewidth}
\centering
\caption{RoboTwin 2.0 success (\%) by step budget: short (S), 400--500; medium (M), 501--900; long (L), 901--1700.}
\label{tab:horizon}
\footnotesize
\setlength{\tabcolsep}{2pt}
\scalebox{0.95}{\begin{tabular}{@{}llrrr|r@{}}
\toprule
Type & Method & S (29) & M (15) & L (6) & All 50 \\
\midrule
\multirow{4}{*}{VLA} & $\pi_0$ & 61.52 & 68.43 & 49.58 & 62.16 \\
 & $\pi_{0.5}$ & 81.31 & 80.60 & 70.08 & 79.75 \\
 & X-VLA & 76.41 & 68.93 & 65.50 & 72.86 \\
 & LingBot-VLA & 85.29 & 87.70 & 84.50 & 85.92 \\
\midrule
\multirow{2}{*}{WAM} & Motus & 89.09 & 85.93 & 86.00 & 87.77 \\
 & Fast-WAM & 93.48 & \textbf{90.53} & 83.67 & 91.42 \\
\midrule
Ours & \method & \textbf{94.00} & 90.27 & \textbf{90.25} & \textbf{92.43} \\
\bottomrule
\end{tabular}
}
\end{minipage}
\end{table}

\paragraph{Memory refresh frequency.}
\label{sec:refresh-ablation}
We compare memory-refresh schedules using the last-layer memory variant. As shown in
Table~\ref{tab:refresh-ablation}, refreshing the memory every action chunk
changes success by $+1.80$ percentage points on standard LIBERO and $-1.23$ percentage points
on LIBERO-Plus relative to once-per-episode construction, while increasing
foundation-model calls by $17.37\times$ and amortized inference time by
$3.12\times$ (3.975 to 12.391\,ms per environment step). Refreshing every two
or four chunks still costs $2.03\times$ and $1.48\times$ the amortized time
while improving standard success by $1.25$ points but reducing LIBERO-Plus
success by $3.07$ and $5.46$ points. These results support
once-per-episode memory construction as an efficient
trade-off between success rate and VLM computation.
\begin{table}[!t]
\centering
\caption{Task-memory refresh ablation on LIBERO using last-layer memory
and matched training for each schedule. Amortized cost includes a 34.25-ms
actor call per chunk and one VLM prefill per memory construction, using the
episode-length distribution and accounting of Table~\ref{tab:libero}.
The once-per-episode cost is 3.975\,ms/step, as in that table.}
\label{tab:refresh-ablation}
\small
\newsavebox{\refreshablationbox}
\newlength{\refreshablationwidth}
\begin{lrbox}{\refreshablationbox}
\begin{tabular}{lrrrr}
\toprule
Memory schedule & VLM calls/episode $\downarrow$ & Amort. ms/step $\downarrow$ & Std. $\uparrow$ & Plus $\uparrow$ \\
\midrule
Every action chunk & 17.37 & 12.391 & \textbf{96.50} & 72.47 \\
Every 2 chunks & 8.93 & 8.052 & 95.95 & 70.63 \\
Every 4 chunks & 4.74 & 5.898 & 95.95 & 68.24 \\
\midrule
Once per episode & \textbf{1.00} & \textbf{3.975} & 94.70 & \textbf{73.70} \\
\bottomrule
\end{tabular}
\end{lrbox}
\global\refreshablationwidth=\wd\refreshablationbox
\usebox{\refreshablationbox}\par
\vspace{2pt}

\end{table}

\paragraph{Memory encoders.}
\label{sec:backbone-ablation}
We construct task memory with four frozen VLMs while keeping the actor
architecture unchanged. Table~\ref{tab:backbone-ablation} shows that all four
provide usable task memory across standard LIBERO and LIBERO-Plus.
Encoder choice has a larger effect on LIBERO-Plus: standard success spans
$94.90$--$95.95\%$, whereas DM0.5 leads the next-best encoder on
LIBERO-Plus by $3.56$ percentage points.
PaliGemma offers a compact alternative with a $9.8$-MB cache versus DM0.5's
$79.7$~MB, but its lower LIBERO-Plus success ($73.15\%$ versus $79.11\%$)
illustrates the trade-off between cache footprint and robustness in this comparison.
\begin{table}[!t]
\centering
\caption{Task-memory encoders on LIBERO (\%). The actor architecture is
unchanged across rows; memory denotes the bf16 KV cache stored per episode
in the LIBERO configuration.}
\label{tab:backbone-ablation}
\fontsize{8}{9}\selectfont
\begin{tabular*}{\refreshablationwidth}{l@{\extracolsep{\fill}}lrrr}
\toprule
Memory producer & Attention & Memory$\downarrow$ & Std. $\uparrow$ & Plus $\uparrow$ \\
\midrule
PaliGemma v1~\citep{paligemma} & 18L, MQA & \textbf{9.8~MB} & 95.35 & 73.15 \\
$\pi_{0.5}$ & 18L, MQA & \textbf{9.8~MB} & 94.90 & 71.48 \\
Qwen2.5-VL-7B~\citep{qwen25vl} & 28L, GQA-4 & 31.8~MB & 95.35 & 75.55 \\
\midrule
DM0.5 (default) & 34L, GQA-4 & 79.7~MB & \textbf{95.95} & \textbf{79.11} \\
\bottomrule
\end{tabular*}

\end{table}

\paragraph{Performance across task horizons.}
\label{sec:horizon}

We evaluate fixed task memory across all 50 RoboTwin 2.0 tasks, grouped by
their official step budgets (Table~\ref{tab:horizon}; evaluation sources are
listed in Appendix~\ref{app:evaluation-details}). For \method, we evaluate
100 episodes per task in each of the Clean and Randomized settings
(5{,}000 episodes per setting). Group success rates pool both settings and
are computed as total successes divided by total episodes; the All 50
column pools all tasks and both settings. For \method, this gives 92.43\%
over 10{,}000 episodes, as reported in Table~\ref{tab:robotwin}. \method achieves 94.00\%,
90.27\%, and 90.25\% success on short-, medium-, and long-horizon tasks,
respectively. On \texttt{put\_bottles\_dustbin}, it reaches 87.00\% success
under a 1{,}700-step budget, which permits up to 34 actor calls using the same
task memory. These results demonstrate effective memory reuse across the
evaluated task horizons.

\section{Conclusion}
\method separates one-time task understanding from repeated closed-loop control. A frozen 4B VLM builds episode-level task memory, while the online actor uses current visual and proprioceptive feedback to produce action chunks. The policy reaches 92.46\%/92.40\% success on RoboTwin Clean/Randomized and 95.95\%/79.11\% on standard LIBERO/LIBERO-Plus. Same-device measurements show 10.9--17.1$\times$ lower RoboTwin amortized inference time than Faster-WAM (HUST), DM0.5, and Fast-WAM. Task-memory swap interventions show that the actor relies on memory matched to the episode.

\label{paper:endmain}

\section*{AI Use Statement}
\label{sec:ai-statement}

The authors used AI tools to help identify and explore related work and to
support research ideation. After drafting the manuscript, the authors also
used AI tools to polish the English writing, including correcting grammar
and improving clarity and phrasing. The authors take full responsibility
for the content of the paper.

\clearpage
\bibliography{refs}
\bibliographystyle{iclr2027_conference}

\clearpage
\appendix
\section*{Appendix}
\section{Architecture and training configuration}
\label{app:configuration}

Table~\ref{tab:architecture-training} summarizes the architecture and training
configuration of the main models on RoboTwin 2.0 and LIBERO.

\begin{table}[H]
\centering
\caption{Architecture and training configuration. Each camera has a separate,
trainable DINOv2 encoder. Batch sizes are effective global batch sizes.}
\label{tab:architecture-training}
\small
\setlength{\tabcolsep}{4pt}
\renewcommand{\arraystretch}{1.15}
\begin{tabular}{@{}p{0.29\linewidth}p{0.335\linewidth}p{0.335\linewidth}@{}}
\toprule
Setting & RoboTwin 2.0 & LIBERO \\
\midrule
Visual encoder & DINOv2 ViT-B/14 & DINOv2 ViT-B/14 \\
Actor image resolution & $168\times224$ & $224\times224$ \\
VLM prefill images & $448\times448$, 3 cameras & $448\times448$, 2 cameras \\
VLM image tokens & 768 & 512 \\
Decoder width / layers / heads & 256 / 4 / 8 & 256 / 4 / 8 \\
Cross-memory GQA KV heads & 8 & 8 \\
History offsets & $[-17,-8,0]$; history from head camera only & $[0]$; no history \\
Action normalization & Z-score & Quantile \\
Demonstrations & 27,500 episodes, 50 tasks; clean + randomized & 2,000 episodes, 40 tasks \\
Effective batch size & 256; 2-way data parallel, 128 per device, accumulation 1 & 32; single device\\
Learning rate & $3\times10^{-4}$ & $3\times10^{-4}$ \\
DINOv2 learning-rate multiplier & 0.1 & 0.1 \\
Learning-rate schedule & 1,000-step warmup + cosine & 1,000-step warmup + cosine \\
Gradient clipping threshold & 1.0 & 1.0 \\
Training precision / loss & BF16 autocast / $\ell_1$ & BF16 autocast / $\ell_1$ \\
Action-chunk tail handling & Repeat & Mask \\
Trainable parameters & 272.0M & 184.7M \\
\bottomrule
\end{tabular}
\end{table}

\section{Latency and memory measurement protocol}
\label{app:measurement-protocol}

Table~\ref{tab:measurement-protocol} specifies the execution settings and
measurement scope used for inference profiling.

\begin{table}[H]
\centering
\caption{Inference latency and memory measurement protocol.}
\label{tab:measurement-protocol}
\small
\setlength{\tabcolsep}{4pt}
\renewcommand{\arraystretch}{1.15}
\begin{tabular}{@{}p{0.23\linewidth}p{0.73\linewidth}@{}}
\toprule
Item & Protocol \\
\midrule

Execution precision & BF16 for the VLM; BF16 autocast with gradient computation disabled for the actor. \\
Batch size & 1. \\
Warmup and timing & 3 warmup calls followed by 10 timed calls; report mean latency. \\
Device synchronization & Synchronized immediately before and after each timed call. \\
Preprocessing and transfer & Timing starts from raw uint8 HWC NumPy images and includes resizing, normalization, and host-to-device transfer. \\
Inference backend & PyTorch 2.7.1+cu128; no {\normalsize\texttt{torch.compile}} or TensorRT. \\
HTTP communication & All methods are invoked directly through their inference functions; HTTP communication is excluded. \\
Memory statistics & For all methods, including DM0.5, PyTorch {\normalsize\texttt{max\_memory\_allocated}} and {\normalsize\texttt{memory\_allocated}} measure peak and current allocated memory, respectively. \\
\bottomrule
\end{tabular}
\end{table}

\section{Additional evaluation details}
\label{app:evaluation-details}

\paragraph{Inference profiling.}
The matched same-hardware measurements cover 10,000 RoboTwin episodes and 2,372,728
executed environment steps, accounting for each method's call schedule.
In the RoboTwin configuration, \method's reusable task memory occupies 174.6\,MiB of the
1.22\,GiB online resident memory after VLM prefill.

\paragraph{Sources of baseline success rates.}
In Table~\ref{tab:robotwin}, the Clean/Randomized results for $\pi_0$,
$\pi_{0.5}$, Motus, and LingBot-VA are taken from Table~1 of
Fast-WAM~\citep{fastwam}; X-VLA's results are from Table~7 of
StarVLA~\citep{starvla}. LingBot-VLA uses the RGB-only (w/o depth)
variant in Table~2 of its original report~\citep{lingbotvla}.
In Table~\ref{tab:libero}, standard LIBERO results marked $^\flat$
($\pi_0$-FAST, OpenVLA-OFT, and UniVLA) come from the
\href{https://github.com/sylvestf/LIBERO-plus/blob/main/libero_res.md}{standard-LIBERO compilation}
maintained by LIBERO-Plus~\citep{liberoplus}; OpenVLA-OFT uses the
third-person-plus-wrist entry. LIBERO-Plus results for $\pi_0$,
$\pi_0$-FAST, OpenVLA-OFT, and UniVLA are taken from the
\href{https://github.com/sylvestf/LIBERO-plus#-libero-plus-benchmark-leaderboard}{official benchmark leaderboard}.
The $\pi_{0.5}$ and Faster-WAM (Noah) LIBERO-Plus results follow
\citet{fasterwamnoah}, while Fast-WAM, Joint-WAM, and Faster-WAM (HUST)
follow \citet{fasterwamhust}. Other baseline success rates follow the
respective model papers or official release reports cited in the tables,
except entries marked $^\parallel$, which are our evaluations of released
checkpoints. Published success rates are retained for Faster-WAM (Noah);
the $^\sharp$ marker applies to our implementation used for cost profiling.
Latency and memory figures are measured separately under our profiling protocol.

\paragraph{Task-horizon evaluation.}
Per-task results for $\pi_0$, $\pi_{0.5}$, X-VLA, Motus, and LingBot-VLA
come from the 50-task data-scaling table in the StarVLA repository's
RoboTwin README~\citep{starvla}, at
\href{https://github.com/starVLA/starVLA/blob/a376315d328b7c47a18497f09ef59ecbc37e37d1/examples/simBenchmarks/Robotwin/README.md}{revision \texttt{a376315}}.
This per-task compilation is distinct from the aggregate table in the
StarVLA paper; we use its LingBot-VLA w/o depth columns. Fast-WAM results come from our local evaluation of its
released checkpoint on 50 tasks, with 100 episodes per task in each of the
Clean and Randomized settings. We pool Clean and Randomized outcomes within each group defined by the
official task step budgets and compute success as total successes divided by
total episodes. With equal episode counts per task and setting, this is also
the mean of the per-task success rates. These per-task evaluations are
sourced separately from the aggregate results in Table~\ref{tab:robotwin}.

\section{Qualitative rollouts}
\label{app:qualitative}

\begin{figure}[H]
\centering
\input{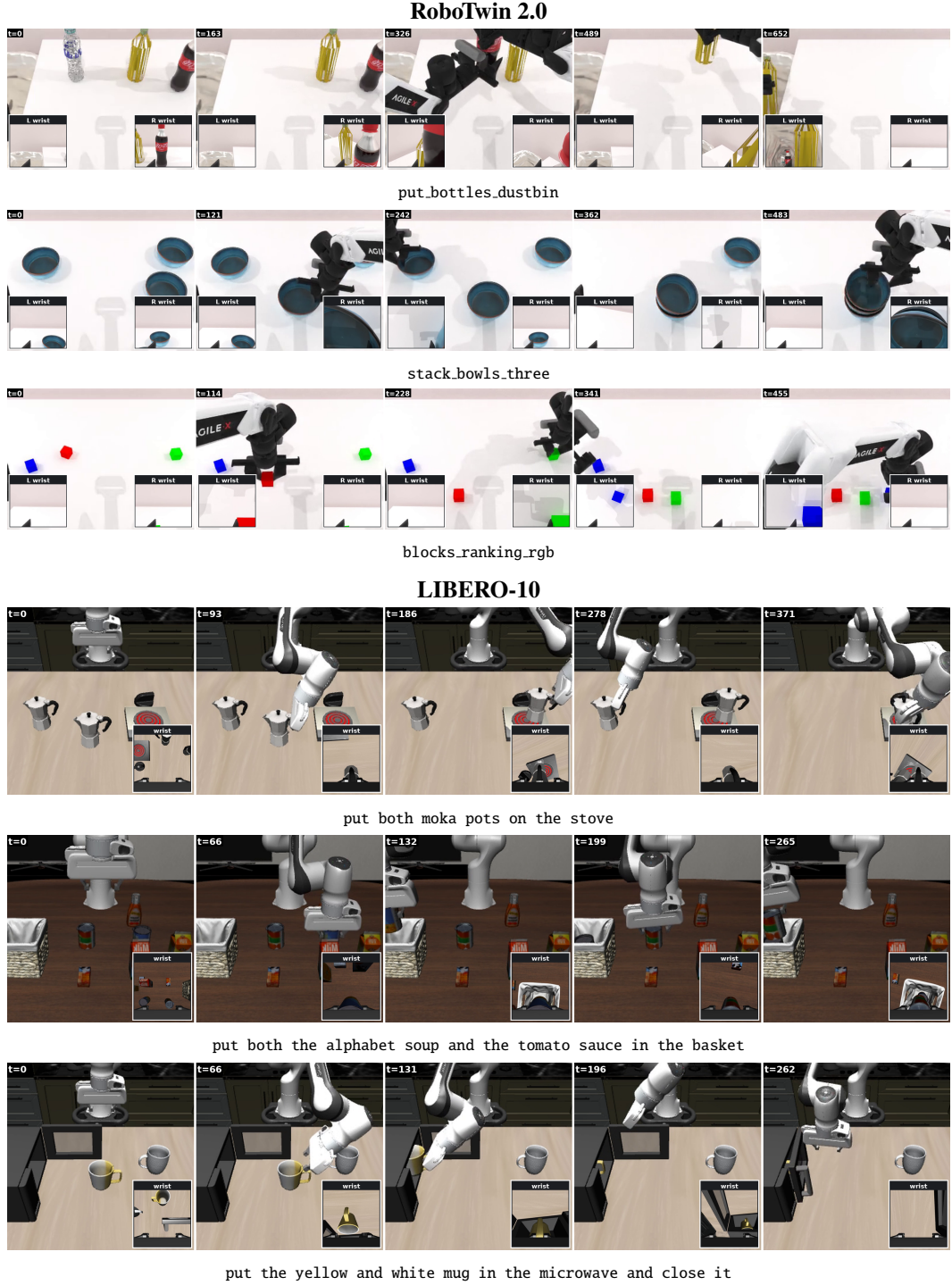}
\caption{Successful rollouts of \method. Each row samples five frames at equal
intervals from one episode; the corner label is the frame index.
Each frame shows the third-person view at full size with the actor's wrist
views inset at the bottom, two for RoboTwin and one for LIBERO. All actions in a row are
produced from a single task memory built before the first frame and never
refreshed.}
\label{fig:rollouts}
\end{figure}

\begin{figure}[p]
\centering
\input{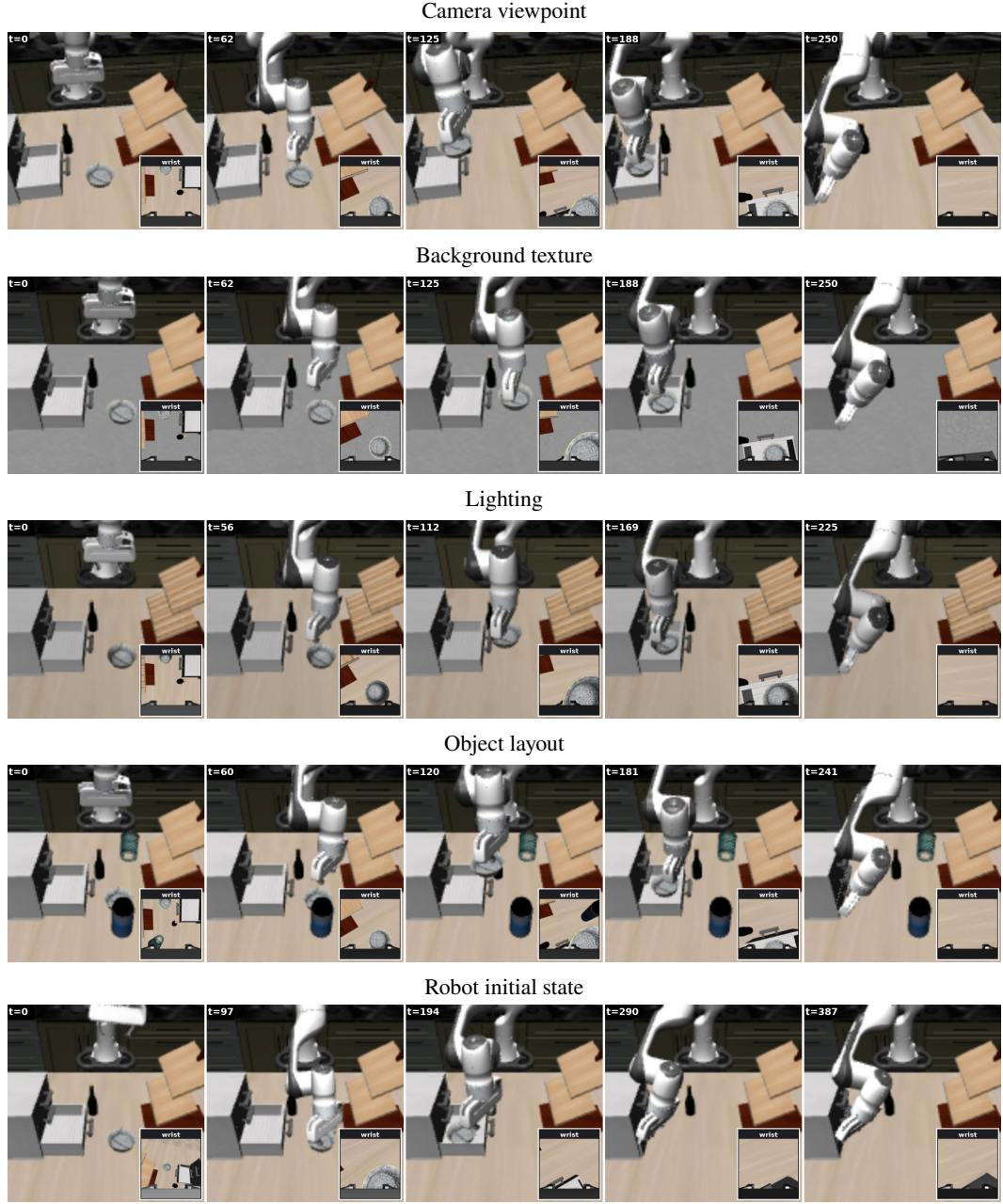}
\caption{Additional successful LIBERO-Plus rollouts for ``put the black bowl
in the bottom drawer of the cabinet and close it.'' Rows show camera,
background, lighting, object-layout, and initial-state perturbations,
respectively. Each row contains five equally spaced frames, with frame
indices in the corners and the wrist view inset. Task memory is built once
from the perturbed initial observations and remains fixed for the episode.
The sensor-noise variant appears in Figure~\ref{fig:rollout-examples}.}
\label{fig:rollouts-plus}
\end{figure}

Figure~\ref{fig:rollouts} shows three successful rollouts from RoboTwin 2.0
and three from LIBERO-10. Figure~\ref{fig:rollouts-plus} complements the
sensor-noise example in Figure~\ref{fig:rollout-examples} with five other
LIBERO-Plus perturbations of the same drawer task: camera viewpoint,
background texture, lighting, object layout, and robot initial state.
Task memory is constructed from the perturbed initial observations and remains
fixed throughout each episode.

\section{Limitations}
\label{app:limitations}

Fast-VLA constructs task memory once at episode initialization and keeps it fixed
throughout execution. While the online actor responds to current observations,
handling changes in task goals or newly introduced instructions during an episode
remains unexplored. Combining Fast-VLA with event-triggered memory updates and
high-level replanning could extend it to such dynamic settings while retaining
the efficiency of memory reuse between updates.

\end{document}